# Fair comparison of skin detection approaches on publicly available datasets

Alessandra Lumini[a*]

[a]*DISI, Università di Bologna, Via dell'Università 50, 47521 Cesena, Italy.*
**correspondinng author*
*EMAIL: alessandra.lumini@unibo.it*

Loris Nanni[b]

[b] *DEI - University of Padova, Via Gradenigo, 6 - 35131- Padova, Italy.*
*EMAIL: loris.nanni@unipd.it*

# Fair comparison of skin detection approaches on publicly available datasets

Alessandra Lumini[a*] and Loris Nanni[b]

[a] *DISI, Università di Bologna, Via dell'Università 50, 47521 Cesena, Italy.*
[b] *DEI - University of Padova, Via Gradenigo, 6 - 35131- Padova, Italy.*

## Abstract

Skin detection is the process of discriminating skin and non-skin regions in a digital image and it is widely used in several applications ranging from hand gesture analysis to track body parts and face detection. Skin detection is a challenging problem which has drawn extensive attention from the research community proposal in the context of expert and intelligent systems, nevertheless a fair comparison among approaches is very difficult due to the lack of a common benchmark and a unified testing protocol. In the recent era, the success of deep convolutional neural network (CNN) has strongly influenced the field of image segmentation detection and gave us various successful models to date. Anyway due to the lack of large ground truth for skin detection only few works have addressed the skin detection problem. In this work, we investigate the most recent researches in this field and we propose a fair comparison among approaches using several different datasets.

The major contributions of this work are (i) an exhaustive literature review of skin color detection approaches and a comparison of approach that can be useful to researcher and practitioners to select the most suitable method for their application, (ii) the collection and examination of many datasets with ground truth for skin detection that can be useful to produce a training set for CNN models, (iii) a framework to evaluate and combine different skin detector approaches, whose source code is made freely available for future research, and (iv) an extensive experimental comparison among several recent methods which have also been used to define an ensemble that works well in many different problems.

Experiments are carried out in 10 different datasets including more than 10000 labelled images: experimental results confirm that the best method here proposed obtains a very good performance with respect to other stand-alone approaches, without requiring ad hoc parameter tuning.

A MATLAB version of the framework for testing and of the methods proposed in this paper will be freely available from https://github.com/LorisNanni.

## 1. Introduction

Skin texture and color are important signs that people use to understand variety of culture-related aspects about each other, as: health, ethnicity, age, beauty, wealth and so on. The presence of skin color in images and videos is a signal of the presence of humans in such media. Therefore, in the last two decades extensive research in the context of expert and intelligent systems has focused on skin detection in videos and images. Skin detection is the process of discriminating "skin" and "non-skin" regions in a digital image and consists in performing a binary classification of pixels and in executing a fine segmentation to define the boundaries of the skin regions. Currently, skin detection is a sophisticated process involving not only the training of models, but also numerous additional methods, including data pre-processing and post-processing.

Skin detection is used within many application domains: it is used as a preliminary step for face detection (Hsu, Abdel-Mottaleb, & Jain, 2002) and tracking (De-La-Torre, Granger, Radtke, Sabourin, & Gorodnichy, 2015), body tracking (Argyros & Lourakis, 2004), hand detection (Roy, Mohanty, & Sahay, 2017) and gesture recognition (Han, Award, Sutherland, & Wu, 2006), biometric authentication (i.e. palm print recognition) (Sang, Ma, & Huang, 2013), objectionable content filtering (Lee, Kuo, Chung, & Chen, 2007), medical imaging. In this work a comprehensive analysis is carried out of how different expert systems (including artificial intelligence, deep learning and machine learning systems) are designed in order to deal with the skin detection problem.

A useful feature for the discrimination of skin and non-skin pixels is the pixel color; nevertheless, obtaining skin color consistency across variations in illumination, diverse ethnicity and different acquisition devices is a very challenging task. Moreover, skin detection, when used as preliminary step of other applications, is required to be computationally efficient, invariant against geometrics transformations, partial occlusions or changes of posture/facial expression, insensitive to complex or pseudo-skin background, robust against the quality of the acquisition device. The factor that worst influences skin detection is the color constancy problem (Khan, Hanbury, Stottinger, & Bais, 2012): i.e. the dependency of pixel intensity on both reflection and illumination which have a nonlinear and unpredictable behavior. To be effective when the illumination conditions vary rapidly, some skin detection approaches use image preprocessing strategies based on color constancy (i.e. a color correction method based on an estimate of the illuminant color) and/or dynamic adaptation techniques (i.e. the transformation of a skin-color model according to the changing illumination conditions). Static skin color approaches that rely on image preprocessing can only partially solve this problem and their performance strongly degrades in real-world applications. A possible solution is considering additional data acquired out the visual spectrum (i.e. infrared images (Kong, Heo, Abidi, Paik, & Abidi, 2005) or spectral imaging (Healey, Prasad, & Tromberg, 2003)), however the use of such sensors is not appropriate for all applications and requires higher acquisition costs which limit their use to specific problems.

Skin detection is a challenging problem and has been extensive studied from the research community. Despite the large number of methods, there are only few surveys in this topic: the works in (C. Prema & Manimegalai, 2012)(Kakumanu, Makrogiannis, & Bourbakis, 2007) are quite old and cover only the methods proposed before 2005, the surveys in (W. Chen, Wang, Jiang, & Li, 2016)(Mahmoodi & Sayedi, 2016)(Naji, Jalab, & Kareem, 2018) are more recent and contain a good investigation of methods, benchmarking datasets and performance related to a period of about two decades. Anyway in none of the above surveys is there a fair comparison among methods using the same testing protocols and datasets. The aim of the present work is not limited to survey the most recent research in this field (which is now enriched of methods based on deep learning (Xu, Zhang, & Wang, 2015)(Zuo, Fan, Blasch, & Ling, 2017)(Kim, Hwang, & Cho, 2017a)(Ma & Shih, 2018)), but also, and above all, to propose a framework for a fair comparison among approaches.

In this research, a novel framework is proposed that integrates different skin color classification approaches and compare their performance and their combination on several publicly available datasets. The major contributions of this research work are:

- An exhaustive literature review of skin color detection approaches with a detailed description of methods freely available.
- The collection and examination of almost all the datasets with ground truth for skin detection available in the literature. Such collection, which includes more than 10000 labelled images, can be useful to produce a training set for CNN models.
- A framework to evaluate and combine different skin detector approaches. The source code of the framework and many of the tested methods will be made freely available for future research and comparisons. The system can be tuned according to the target application: on the basis of the application requirement, the acceptance threshold can be tuned to prune a large percentage of false accepts at a small cost of reduction in genuine accepts or vice versa a larger number of false accepts can be admitted to maximize the number of genuine accepts. The framework includes training and testing protocols for most used benchmark datasets in this field.
- A fair comparison among the most recent research and methods in the skin detection field, using the same testing protocols, benchmark datasets and performance indicators. An evaluation of computation time of each method in order to perform a comparison also in terms of complexity. A discussion about performance can help researchers and practitioners in evaluating the approaches most suited to their requirements according to computational complexity, memory requirements, detection rate and sensitivity.
- Three different CNN architectures are trained for skin detection and the model are made available.

The arrangement of this paper is as follows. In section 2 related works in skin detection are presented, including a discussion about taxonomy of existing approaches and a detailed description of the approaches tested in this work. In section 3 the evaluation problem is treated, the most known datasets used for performance evaluation are listed and commented, testing protocols and performance indicators used in our experiment are discussed. In section 4 the experiments conducted using the proposed framework are reported and discussed. Finally, section 5 includes the conclusions and some future research directions.

## 2. Skin detection approaches

Several skin detection methods are based on the assumption that skin color can be recognized from background colors according to some clustering rule in a specific color space. Even if this assumption can be valid in a constrained environment where both ethnicity of the people and background colors are known, it is a very challenging task in complex images captured under unconstrained conditions and when individuals show a large spectrum of human skin coloration (Kakumanu et al., 2007). There are a lot of challenging factors that influence the performance of a skin detector:

- Human characteristics as ethnicity and age: skin color ranges from white to dark brown among human racial groups, the transition from fresh skin to dry skin related to the age determines a strong variation of tones.
- Acquisition conditions: factors such as camera characteristics or illumination variations strongly influence the skin appearance. Generally, a variation in the illumination level or in the light source distribution determines the presence of shadows and changes in the color of the skin.
- Skin painting: the presence of makeup or tattoo coverage affects the appearance of skin.
- Complex background: sometimes the skin detector can be deceived by the presence of objects that have skin-like color in the background.

Skin detection approaches can be classified according to several taxonomy schemes which evaluate different aspects of the methods:

1. Considering the presence or not of preprocessing steps such as color correction and illumination cancelation (Zarit, Super, & Quek, 1999) or dynamic adaptation (Ibrahim, Selim, & Zayed, 2012) designed to reduce the effects of different acquisition conditions;
2. Considering the color space used for pixel classification (Kakumanu et al., 2007). Most of research papers dealing with the skin detection have to face the problem of selecting the most appropriate skin color model (Naji et al., 2018). The performance of several different color models are compared in (Kakumanu et al., 2007): basic models (i.e. RGB, normalized RGB), perceptual models (i.e. HIS, HSV) perceptual uniform models (i.e. CIE-Lab, CIE-Luv) and orthogonal models (i.e. YCbCr, YIQ) with the finding that orthogonal models are characterized by a reduced redundancy/correlation among channels, therefore they are the most suited for skin detection.
3. Examining the problem formulation, which can be based on the segmentation of an image in the regions where human skin is present (segmentation based approaches) or on treating each pixel as skin or non-skin without considering its neighbors (pixel-based approaches). Despite the presence of a huge number of pixel-based approaches (Kakumanu et al., 2007), the region-based skin color detection techniques are very few (Poudel, Zhang, Liu, & Nait-Charif, 2013)(W. C. Chen & Wang, 2007)(Kruppa, Bauer, & Schiele, 2002)(Sebe, Cohen, Huang, & Gevers, 2004). Some recent methods (Zuo et al., 2017)(Kim, Hwang, & Cho, 2017b) based on convolutional neural networks can be included in this category.

4. Distinguish among methods for performing pixel classification (i.e. explicitly defined skin region, parametric approaches, non-parametric approaches) (Kumar & Malhotra, 2015). The first category, also named rule based, defines explicit rules to define the skin color (in an opportune color space), The other categories include machine learning approaches that make use of non-parametric or parametric approaches to estimate the color distribution from a training set (returning a lookup table or a parametric model, respectively).

5. Considering the type of classifier used for machine learning approaches: in (Mahmoodi & Sayedi, 2016) a taxonomy of 8 not exclusive groups is proposed which extends the simple division in parametric and non-parametric approaches. *Statistical methods* include parametric methods based on Bayesian rule o Mixture models (Jones & Rehg, 2002). *Neural networks models* (QingXiang Wu, Cai, Fan, Ruan, & Leng, 2012)(L. Chen, Zhou, Liu, Chen, & Xiong, 2002) are used for the segmentation of color images based on both color and texture information. *Diffusion based methods* (Mahmoodi & Sayedi, 2014)(Nidhu & Thomas, 2014) extend analysis on neighboring pixels to improve the classification performance: after an initial pixel based extraction a seed growing method is applied to include neighborhood in the skin region. *Adaptive methods* (Kawulok, Kawulok, Nalepa, & Smolka, 2014) are based on the tuning of models in order to adapt to specific conditions (i.e. lighting, skin tone, background); the model calibration often grants performance advantages, but at the cost of increased computation time. *Hyperspectral models* (Nunez & Mendenhall, 2008) are based on acquisition devices with hyperspectral capabilities: despite the advantage due to the presence of spectral information, these approaches will be not considered in this work since they are only applicable to specifically collected datasets. *SVM based systems* are parametric models based on SVM classifier: this class also overlaps adaptive methods if the SVM classifier is trained using active learning (Han et al., 2006). *Mixture techniques* are the methods based on the mixture of different machine learning approaches (Jiang, Yao, & Jiang, 2007)(Al-Mohair, Mohamad Saleh, & Suandi, 2015).

In the last few years the research in skin detection has taken two main directions, according to the following consideration: even if in the most general applications nothing should be assumed about the background and the acquisition conditions, in many applications the difference between skin and background is large, the acquisition conditions are controlled and the skin region is quite easy to detect. For example, in many gesture recognition applications the hand images are acquired using flatbed scanners and have dark unsaturated backgrounds (Sandnes, Neyse, & Huang, 2016). For this reason, in addition to many approaches that adopt sophisticated and computationally *expensive techniques*, several *simple rule-based methods* have been proposed, which are preferred in some applications since they are easier to understand, implement and reuse, more efficient, while at the same time, adequately effective. Usually *simple rule-based methods* are not even tested on benchmarks for pure skin detection but as a step of a more complex task (i.e. face detection, hand gesture recognition). A recent example of method belonging to this class is the work in (Sandnes et al., 2016), which performs a study on different color models, drawing the conclusion that there are no apparent advantages of using a perceptual uniform color space: therefore they propose an approach based on a simple RGB lookup table. The work in (Jairath, Bharadwaj, Vatsa, & Singh, 2016) proposes a dynamic but very straightforward method based on parametric modeling the Cr or Cb channel from YCbCr color space through a single Gaussian: the final approach requires a limited amount of storage space, is fast and can be trained using a small training set with small time delay. In (Song, Wu, Xi, Park, & Cho, 2016) a simple skin detector based on RGB histogram thresholding is employed as a preliminary fast step for motion-based skin region of interest detection. A dynamic skin color model method based on space switching is proposed in (Gupta & Chaudhary, 2016), where in order to handle with natural changes in illumination, a system with three robust algorithms has been built, based on different color spaces which are switched according to the statistical mean of the skin pixels in the image. In (Oghaz, Maarof, Zainal, Rohani, & Yaghoubyan, 2015) a new 3D hybrid color space named SKN is proposed, obtained by using Principal Component Analysis and a Genetic Algorithm to discover the optimal representation of skin color over 17 color spaces, then a pixel-wise skin classification is performed employing a general purpose trained classifier (i.e. Naïve Bayes, Random Forest, Multilayer Perceptron and Support Vector Machines).

The first class of methods, based on sophisticated and computationally *expensive techniques*, include recently proposed approaches based on deep learning (Xu et al., 2015)(Zuo et al., 2017)(Kim et al., 2017b)(Ma & Shih, 2018). Convolutional neural networks have recently achieved remarkable results for a variety of computer vision tasks, including several applications based on pixel-wise prediction (i.e. scene labelling, semantic image segmentation). In (Xu et al., 2015) a patch-wise skin segmentation approach is described based on deep neural networks, which uses image patches as processing units instead of pixels; a dataset of image patches have been appositely collected for training purposes and the trained deep skin models are integrated into a sliding window framework to detect skin regions of the human body parts achieving competitive performance on pixel-wise skin detection. In (Zuo et al., 2017) an integration of some recurrent neural networks layers into the fully convolutional neural networks is proposed as a solution to develop an end-to-end network for human skin detection. In (Kim et al., 2017b) the authors propose a inception-like convolutional network-in-network structure, which consists of convolution and rectified linear unit layers only (without pooling and subsampling layers), while in (Ma & Shih, 2018) the authors make experiments on many CNN structures to determine the best one for skin detection. All the proposed architectures report comparable or better performance than other the state-of-the-art methods; anyway, a fair comparison with traditional existing approaches is always difficult due to the different testing protocols. For example in a very recent work (Dourado, Guth, de Campos, & Li, 2019), the authors suggest to compare some deep learning approaches on four different datasets giving both in-domain and cross-domain results: for the latter approach they report different performance depending on the training set. Since none of the authors above has released a model to perform comparisons, in this work we train and use three of the most recently proposed models for image segmentation: SegNet (Badrinarayanan, Kendall, & Cipolla, 2017), U-Net (Ronneberger, Fischer, & Brox, 2015) and Deeplabv3+ (L. C. Chen, Zhu, Papandreou, Schroff, & Adam, 2018). SegNet is a convolutional encoder-decoder net: it uses VGG16 or VGG19 with only forward connections and non trainable layers as encoder, then for each of the encoder layers there is a corresponding decoder which upsamples the feature map using max-pooling indices. The U-Net architecture is a combination of convolutions and transposed convolutions layers used downsampling and then upsampling the images spatially. The U-Net architecture has reported impressive results in medical image segmentation. DeepLabv3+ is based on spatial pyramid pooling module and encode-decoder structure: it is built on top of a powerful convolutional neural network (CNN) backbone architecture thus it can exploit performing pre-trained networks as decoders.

In this work, we evaluate and combine several skin detectors with the aim of comparing their performance and we propose an ensemble able of maximizing their classification performance. The base approaches used to create the ensemble have been selected according to their effectiveness, their availability to research scopes and their efficiency:

- *GMM* (Jones & Rehg, 2002) is a simple and efficient Gaussian mixture model for skin detection trained to classify non-skin vs. skin pixels in the RGB space.
- *Bayes* (Jones & Rehg, 2002) is a fast and effective method based on a Bayesian classifier. In this work we used a classifier trained in the RGB color space using 2000 images from the ECU data set.
- *SPL*($\tau$) (Conaire, O'Connor, & Smeaton, 2007)[1] is a pixel-based skin detection approach which determines skin probability in the RGB domain using a look up table (LUT). For a testing image the probability of being skin for each pixel $x$ is calculated, then a threshold $\tau$ is applied to decide *skin*/*non-skin*.
- *Cheddad*($\tau$) (Cheddad, Condell, Curran, & Mc Kevitt, 2009) is a pixel-based and real-time approach which reduces the RGB color space to a 1D space derived from differentiating the grayscale map and the non-red encoded grayscale version. The classification is performed using a skin probability which delimits the lower and upper bounds of the skin cluster and the final decision depends on a classification threshold $\tau$;
- *Chen* (Y. H. Chen, Hu, & Ruan, 2012) is a statistical skin color model, which is specifically suited to be implemented on hardware. The 3D skin cube is represented as three 2D sides calculated as the difference of two color channels: sR=R-G, sG=G-B, sB=R-B. The skin cluster region is delineated in the transformed space, fixing the boundaries to the following ranges: $-142<sR<18$, $-48< sG <92$, and $-32< sB <192$.
- *SA1*($\tau$) (Kawulok, 2013) is a method for skin detection based on spatial analysis. Starting from a skin probability map obtained using a color pixel-wise detector, the starting step to the spatial analysis is the proper selection of the high probability pixels as "skin seeds". The second step consists in finding the shortest routes from each seed to every single pixel, in order to propagate the "skinness". All the pixels not adjoined during the propagation process are labeled as non-skin. The performance depends on the threshold $\tau$ used to classify.
- *SA2*($\tau$) (Kawulok, Kawulok, & Nalepa, 2014) is another recent method based on spatial analysis which uses both color and textural features to determine the presence of skin. The basic idea is to extract the textural features from the skin probability maps rather than from the luminance channel: therefore simple textural statistics are computed from each pixel's neighborhood in the probability map using kernels of different sizes (i.e. the median, the minimal values, the difference between the maximum and minimum and the standard deviation). Then skin and non-skin pixels are transformed into two classes of feature vectors whose size is reduced by Linear Discriminant Analysis (LDA) to increase their discriminating power. Finally the spatial analysis method proposed in (Kawulok, 2013) and described above is used for seed extraction and propagation using the distance transform. The classification depends on a threshold $\tau$ used to classify pixels in the distance domain.
- *SA3*($\tau$) (Kawulok, Kawulok, Nalepa, et al., 2014) is a self-adaptive method that consists in combining a local skin color model created using a probability map and spatial analysis to fix the boundaries of the skin regions. It is an evolution of the method proposed in (Kawulok, Kawulok, & Nalepa, 2014) based on spatial analysis, skin color model adaptation and textural features. The main difference from the method in (Kawulok, Kawulok, & Nalepa, 2014) is a new technique for extracting adaptive seeds, based on the analysis of the skin probability map calculated from the input color image. The performance depends on a classification threshold $\tau$.
- *DYC* (Brancati, De Pietro, Frucci, & Gallo, 2017)[2] is a skin detection approach which works in the YCbCr color space and takes into account the lighting conditions. The method is based on the dynamic generation of the skin cluster range both in the YCb and YCr subspaces of YCbCr and on the definition of correlation rules between the skin color clusters.
- *SegNet* is a deep learned approach for image segmentation. In our experiments the tested architecture has been modeled starting from a pre-trained VGG19 network and then fine-tuned using the first 2000 images of the ECU dataset according to the following training options: batch size 32, learning rate 0.001, max epoch 30 (data augmentation, i.e. flip and small rescaling, has been used only in the first 15 epoch). Moreover, we use inverse class frequency weighting in order to deal with imbalanced classes: the frequencies of "skin" and 'non-skin' labels over the training data are used as 'ClassWeights' in final pixelClassificationLayer. This and the following two deep learned models are tested in all the other datasets without further tuning.
- *U-Net* is another encoder-decoder architecture for image segmentation. In our experiments a U-Net with encoder depth 4 has been trained from scratch using the first 2000 images of the ECU dataset and the following options: batch size 32, learning rate 0.001, max epoch 50 (data augmentation has been used only in the first 15 epoch), class weighting.
- *DeepLab* is a DeepLabv3+ network modeled starting from a pre-trained ResNet50. The segmentation model is fine-tuned using the first 2000 images of the ECU dataset according to the following options: batch size 32, learning rate 0.001, max epoch 30 (data augmentation has been used only in the first 15 epoch), class weighting.

A rough classification of the methods tested in this work according to the classification criteria surveyed in section 2 is reported in Table 1.

---

[1] http://clickdamage.com/sourcecode/index.php

[2] https://github.com/nadiabrancati/skin_detection/

**Table 1. Rough classification of the tested approaches**

| | *GMM* | *Bayes* | *SPL* | *Cheddad* | *Chen* | *SA1* | *SA2* | *SA3* | *DYC* | *SegNet* | *U-Net* | *DeelLabv3+* |
|---|---|---|---|---|---|---|---|---|---|---|---|---|
| **Preprocessing steps** | | | | | | | | | | | | |
| None | x | x | x | x | x | | | | | x | x | x |
| Dynamic adaptation | | | | | | x | x | x | x | | | |
| **Color space** | | | | | | | | | | | | |
| Basic color spaces | x | x | x | | | | | | | x | x | x |
| Perceptual color spaces | | | | | | x | x | x | | | | |
| Orthogonal color spaces | | | | | | | | | x | | | |
| Other (e.g. Color ratio) | | | | x | x | | | | | | | |
| **Problem formulation** | | | | | | | | | | | | |
| Segmentation based | | | | | | x | x | x | | x | x | x |
| Pixel based | x | x | x | x | x | | | | x | | | |
| **Type of pixel classification** | | | | | | | | | | | | |
| Rule based | | | | x | x | | | | x | | | |
| Machine learning: parametric | x | x | | | | | | | | | | |
| Machine learning: non-parametric | | | x | | | | | | | | | |
| **Type of classifier** | | | | | | | | | | | | |
| Statistical | | x | x | | | | | | | | | |
| Mixture techniques | x | | | | | | | | | | | |
| Adaptive methods | | | | | | x | x | x | | | | |
| CNN | | | | | | | | | | x | x | x |

## 3. Skin detection evaluation: Datasets and performance indicators

To assist research in the area of skin detection, there are some well-known color image datasets provided with ground truth. The use of a standard and representative benchmark is essential to execute a fair empirical evaluation of skin detection techniques.

### *3.1. Datasets*

In Table 2 some of the most used datasets are summarized and in this section a brief description of each of them is given.

**Compaq** (Jones & Rehg, 2002) is one of the first and most used large skin dataset, it consists of images collected by crawling Web: 9731 images containing skin pixels (but only 4675 skin images have been segmented and included in the ground truth) and 8965 images with no skin pixels. This dataset have been extensity used for testing and comparing methods, but without using a standard testing protocol, therefore comparisons using this dataset are not always fair. Moreover, the ground truth for this dataset has been obtained using an automatic software tool leading to imprecise results.

**TDSD** (Zhu, Wu, Cheng, & Wu, 2004) is an old dataset containing 555 images with very imprecise annotations (automatic labeling).

The **UChile** (Ruiz-Del-Solar & Verschae, 2004) dbskin2 complete set includes 103 images acquired in different lighting conditions and with complex background. The ground truth has been manually annotated with medium precision (in some images the boundaries between skin and non-skin pixels are not marked precisely).

**ECU** (Phung, Bouzerdoum, & Chai, 2005) skin and face datasets are a collection of about 4000 color images annotated with relatively accurate ground truth. ECU dataset is quite challenging, since it includes diversity in terms of the lighting conditions, background scenes, and skin types.

The skin dataset named **Schmugge** (Schmugge, Jayaram, Shin, & Tsap, 2007) is a collection of 845 images taken from different face datasets (i.e. the UOPB dataset, the AR face dataset, and University of Chile database). The ground truth is very accurate since all images are labelled in 3 classes: skin/ not-skin/don't care.

**Feeval** (Stöttinger, Hanbury, Liensberger, & Khan, 2009) is a dataset based on 8991 frames extracted from 25 online videos of low quality. The ground truth is not precisely annotated. Here, the performance indicator is calculated averaging the performance of each video by the number of frames (therefore considering each video as a single image in the performance evaluation).

**MCG** skin database [29] contains 1000 images selected from internet in order to include confusing backgrounds, variable ambient lights and diversity of human races. The ground-truth is obtained through manually labeling, but it is not very precise since eyes, eyebrows, and even bracelets are sometimes marked as skin.

**VMD** (Sanmiguel & Suja, 2013) is obtained selecting 285 images from several public datasets for human activity recognition (i.e. LIRIS, EDds, UT, AMI and SSG). The dataset is already divided in subsets for training and testing, anyway in this work we used all the images for testing. The images cover a wide range of illumination levels and situations.

**SFA** dataset (Casati, Moraes, & Rodrigues, 2013) includes images from FERET (876 images) and AR (242 images) face databases manually labelled (with medium precision). SFA includes an internal organization in folders in order to separate 1118 original images (ORI), 1118 ground truths (GT) masks, 3354 samples of skin (SKIN) and 5590 samples of nonskin (NS) ranging differently from 1 to 35×35 dimensions. We have used ORI/GT for assessing the performance.

**Pratheepan** (Tan, Chan, Yogarajah, & Condell, 2012) is a small dataset which includes 78 images downloaded randomly from Google; the dataset is divided in two subsets: FacePhoto including 32 single subject images with simple background and FamilyPhoto including 46 images with multiple subjects and complex background.

**HGR** (Kawulok, Kawulok, Nalepa, et al., 2014) is a dataset for gesture recognition which contains also ground truth binary skin presence masks; the dataset includes 1558 images representing gestures from Polish and American sign language with controlled and uncontrolled background divided in 3 subsets (HGR1, HGR2A, HGR2B). In our tests the size of the images of HGR2A and HGR2B has been reduced of a factor 0.3.

**SDD** (Mahmoodi et al., 2015) is a dataset of 21000 images acquired using different imaging devices, in a variety of illumination conditions and including different skin colors from people all around the world. The dataset is composed from some images extracted from videos and others belonging to popular face datasets. Images are provided in four sets: a set including mainly single face images, made for training purposes, and other three sets to be used for testing.

**FvNF** (Nanni & Lumini, 2017) (Face vs. NonFace) is not a real skin dataset, it is composed by 800 face and 770 non-face images, extracted from Caltech dataset (Angelova, Abu-Mostafa, & Perona, 2005). This dataset has been collected and used in (Nanni & Lumini, 2017) to evaluate the capability of a skin detector method to detect the presence of a face, based on the number of pixels classified as skin. The average precision ($AP$) is used as performance indicator, $AP \in [0,100]$. The $AP$ summarizes the shape of the precision/recall curve, since it is the area under the precision-recall curve.

**Table 2. Some of the most used datasets per skin detection**

| Name | Ref | Images | Ground truth | Download | Year |
|---|---|---|---|---|---|
| Compaq | (Jones & Rehg, 2002) | 4675 | Semi-supervised | ask to the authors | 2002 |
| TDSD | (Zhu et al., 2004) | 555 | Imprecise | http://lbmedia.ece.ucsb.edu/research/skin/skin.htm | 2004 |
| UChile | (Ruiz-Del-Solar & Verschae, 2004) | 103 | Medium Precision | http://agami.die.uchile.cl/skindiff/ | 2004 |
| ECU | (Phung et al., 2005) | 4000 | Precise | http://www.uow.edu.au/~phung/download.html (currently not available) | 2005 |
| Schmugge | (Schmugge et al., 2007) | 845 | Precise (3 classes) | https://www.researchgate.net/publication/257620282_skin_image_Data_set_with_ground_truth | 2007 |
| Feeval | (Stöttinger et al., 2009) | 8991 | Low quality, imprecise | http://www.feeval.org/Data-sets/Skin_Colors.html | 2009 |
| MCG | (Huang, Xia, Zhang, & Lin, 2011) | 1000 | Imprecise | http://mcg.ict.ac.cn/result_data_02mcg_skin.html (ask to authors ) | 2011 |
| Pratheepan | (Tan et al., 2012) | 78 | Precise | http://web.fsktm.um.edu.my/~cschan/downloads_skin_dataset.html | 2012 |
| VDM | (Sanmiguel & Suja, 2013) | 285 | Precise | http://www-vpu.eps.uam.es/publications/SkinDetDM/ | 2013 |
| SFA | (Casati et al., 2013) | 1118 | Medium Precision | http://www1.sel.eesc.usp.br/sfa/ | 2013 |
| HGR | (Kawulok, Kawulok, Nalepa, et al., 2014) | 1558 | Precise | http://sun.aei.polsl.pl/~mkawulok/gestures/ | 2014 |
| SDD | (Mahmoodi et al., 2015) | 21000 | Precise | Not available | 2015 |

### *3.2. Performance indicators*

Skin detection is a two-class classification problem, therefore standard measures for general classification problems (Powers, 2011) can be used for performance evaluation: including Accuracy, Precision, Recall, $F_1$ measure, Kappa, ROC-Curve, Area Under Curve, and others. Anyway, due to the particular nature of this problem which is based on pixel-level classification and on unbalanced distribution, the following measures are usually considered for performance evaluation: the confusion matrix, the $F_1$ measure, the True Positive Rate ($TPR$) and the False Positive Rate ($FPR$).

The *confusion matrix* is obtained comparing results with the ground-truth data to determine the number of true negatives ($tn$), false negatives ($fn$), true positives ($tp$) and false positives ($fp$). Several useful measures can be obtained from the confusion matrix, including $precision = tp/(tp+fp)$, that is the percentage of correctly classified pixels out of all the pixels classified as skin and $recall = tp/(fn+tp)$, that is the percentage of the ground-truth skin pixels correctly classified as skin.

The $F_1$ *measure* is the harmonic mean of precision and recall and it is calculated according to the following formula $F_1 = 2tp/(2tp + fn + fp)$ , $F_1 \in [0,1]$. According to other works in the literature $F_1$ is averaged at pixel level not at image level; in such a way the final indicator is not dependent on the image size in the different databases. Besides $F_1$ several papers use the *True Positive Rate* (*TPR= recall=tp*/(*fn+tp*)) and the *False Positive Rate* (*FPR= fp*/(*fp+tn*)).

Even if the $F_1$ measure is able to evaluate algorithms only at a fixed operating threshold, and therefore it is a worse way to quantify the accuracy of an algorithm evaluation with respect to ROC curve or precision-recall curve, we use $F_1$ in this work since it is widely used in the literature for skin classification and allow a better comparison. Moreover, several methods evaluated here use a fixed threshold and do not allow tuning for true positives and false positives rates.

## 4. A fair experimental comparison

A fair comparison among different approaches is very difficult due to the lack of a universal standard in evaluation: most of published works are tested on self-collected datasets which often are not available for further comparison; in many cases the testing protocol is not clearly explained, many datasets are not of high quality and the precision of the ground truth is questionable since sometimes lips, mouth, rings and bracelets have been labelled as skin. In this section, we carry out a comparison of some well-known approaches whose source code is freely available. In order to accomplish a fair comparison, we do not perform re-training of methods in each dataset, conversely, we use the knowledge provided by the original authors, therefore all the datasets have been used only for testing purposes. When the performance depends on a classification threshold, we tested several values in order to select the one that performs best in all the datasets.

In Table 3 we report the performance in 10 datasets of the 12 stand-alone approaches described in section 2 and four ensembles combined by the vote rule. Both the stand-alone approaches and the ensembles have been tested using different thresholds (only the best one is reported in Table 3 for sake of space, the complete version of the table is included as supplementary material):

- *GMM* (Jones & Rehg, 2002), *Bayes* (Jones & Rehg, 2002) ($\tau\in$[50,70,90,110,140]), *SPL* (Conaire et al., 2007) ($\tau\in$[-2.5,-2,-1.5,-1,-0.5]), *Cheddad* (Cheddad et al., 2009), *Chen* (Y. H. Chen et al., 2012), *SA1* (Kawulok, 2013) ($\tau\in$[100,150,175,200,225]), *SA2* (Kawulok, Kawulok, & Nalepa, 2014) ($\tau\in$[30,40,50,85,120]), *SA3* (Kawulok, Kawulok, Nalepa, et al., 2014) ($\tau\in$[25,50,75,100,125]), *DYC* (Brancati et al., 2017)

- *SegNet, U-Net* and *DeepLab* are the method based on CNN trained on an external dataset according to the procedure explained in section 2.

- *Vote1*, *Vote2*, *Vote3* and *Vote4* are the ensembles proposed in this work and based on the weighed fusion of the following methods using the vote rule: $w_1\times SA1(175) + w_2\times SA2(50) + w_3\times SA3(50) + w_4\times Cheddad(125) + w_5\times DYC + w_6\times Bayes(110) + w_7\times SegNet + w_8\times U\text{-}Net + w_9\times DeepLab > (w_1+w_2+w_3+w_4+w_5+w_6+w_7+w_8+w_9)/w_\tau$, where $\mathbf{w}=(w_1, w_2, w_3, w_4, w_5, w_6, w_7, w_8, w_9)$ is the weight vector and $w_\tau\in$[1.25,1.5,1.75,2] is a threshold parameter used to tune the sensitivity of the method. The weight arrays used for the four ensembles are, respectively:

  **w1**=(0.5, 1.5, 1, 1.5, 0.5, 1, 0, 0, 0),
  **w2**=(0.5, 1.5, 1, 1.5, 0, 1, 5.5, 0, 0),
  **w3**=(0.5, 1.5, 1, 1.5, 0, 1, 5.5, 2.75, 0);
  **w4**=(0.25, 0.75, 0.5, 0.75, 0, 0.5, 2.75, 1.375, 5.5);

  therefore the first ensemble (*Vote1*) does not include any CNN classifier, *Vote2* includes *SegNet* (weighted as the sum of the other classifiers), *Vote3* includes *SegNet* and *U-Net* (weighed half of *SegNet*), *Vote4* includes all the deep learning methods. In our experiments we have tested different rules for the fusion of classifiers, finding that the vote rule is the best suited for this particular problem.

For each dataset, the best result is highlighted and in the last column the rank (calculated as the rank of the average rank) is reported. Maybe some approaches can be advantaged in some datasets, in case that they used some images of the dataset for training (i.e. LUT creation), anyway considering the results on all the 10 datasets, the comparison is quite fair.

**Table 3. Performance in 10 different datasets**

| | | Dataset | | | | | | | | | | |
|---|---|---|---|---|---|---|---|---|---|---|---|---|
| *Perf. Indicator* | | *$F_1$ measure* | | | | | | | | | *AP* | *Rank* |
| Method | $\tau$/ $w_\tau$ | *Feeval* | *Pratheepan* | *MCG* | *UChile* | *Compaq* | *SFA* | *HGR* | *Schmugge* | *VMD* | *FnF* | |
| *GMM* | | 0.513 | 0.581 | 0.688 | 0.615 | 0.600 | 0.789 | 0.658 | 0.595 | 0.130 | 91.590 | 13 |
| Bayes | 110 | 0.565 | 0.631 | 0.694 | 0.661 | 0.599 | 0.760 | 0.871 | 0.569 | 0.252 | 92.380 | 14.5 |
| *SPL* | -2 | 0.544 | 0.551 | 0.621 | 0.568 | 0.494 | 0.700 | 0.845 | 0.490 | 0.321 | 89.380 | 9 |
| *Cheddad* | 125 | 0.596 | 0.597 | 0.667 | 0.649 | 0.588 | 0.683 | 0.767 | 0.571 | 0.261 | 89.200 | 12 |
| *Chen* | | 0.287 | 0.540 | 0.656 | 0.598 | 0.549 | 0.791 | 0.732 | 0.571 | 0.165 | 87.200 | 16 |
| *SA1* | 175 | 0.558 | 0.613 | 0.664 | 0.567 | 0.593 | 0.788 | 0.768 | 0.482 | 0.199 | 86.770 | 14.5 |
| *SA2* | 50 | 0.515 | 0.693 | 0.755 | 0.663 | 0.645 | 0.771 | 0.806 | 0.594 | 0.156 | 91.910 | 10 |
| *SA3* | 50 | 0.539 | 0.709 | 0.762 | 0.625 | 0.647 | 0.863 | 0.877 | 0.586 | 0.147 | 89.800 | 8 |
| *DYC* | | 0.588 | 0.599 | 0.680 | 0.663 | 0.618 | 0.569 | 0.616 | 0.613 | 0.275 | 88.710 | 11 |
| *SegNet* | | 0.717 | 0.730 | 0.813 | 0.802 | 0.737 | 0.889 | 0.869 | 0.708 | 0.328 | 99.978 | 5 |
| *U-Net* | | 0.576 | 0.787 | 0.779 | 0.713 | 0.686 | 0.848 | 0.836 | 0.671 | 0.332 | 98.848 | 6 |
| *DeepLab* | | 0.771 | 0.875 | **0.879** | **0.899** | 0.817 | 0.939 | 0.954 | 0.774 | **0.628** | **99.995** | 2 |
| *Vote1* | 1.5 | 0.594 | 0.717 | 0.754 | 0.670 | 0.666 | 0.737 | 0.849 | 0.625 | 0.269 | 92.980 | 7 |
| *Vote2* | 1.75 | **0.736** | 0.811 | 0.816 | 0.81 | 0.772 | 0.854 | 0.949 | 0.700 | 0.481 | 99.780 | 4 |
| *Vote3* | 1.75 | 0.733 | 0.812 | 0.841 | 0.829 | 0.773 | 0.902 | 0.950 | 0.714 | 0.423 | 99.800 | 3 |
| *Vote4* | 1.75 | **0.776** | **0.879** | 0.878 | 0.897 | **0.819** | **0.944** | **0.967** | **0.776** | 0.62 | **99.995** | 1 |

From the results in Table 3 it is clear the deep learning revolution: the three CNNs outperform all the previous handcrafted stand-alone approaches. No handcrafted approach overcomes the others in all the problems: it is very difficult to find a solution that works well in all the datasets without retraining and performing a parameter tuning. On the contrary the best performing segmentation network, i.e. DeepLabv3+, gives valuable performance in all the problems, even if trained on a small dataset. The design of an ensemble method able to exploit the good performance of each of its composing approaches is a feasible solution: our proposed ensembles, based on the vote rule, grant good performance on average in almost all the datasets and reach the first positions in the rank column. Anyway the computational overhead of a fusion approach does not make this solution suitable in a real application, mainly considering the good performance of deep learning. In conclusion the performance obtained on 10 different datasets by DeepLabv3+ (without ad hoc training or tuning) is a very strong result compared to other existing approaches and makes this method the best choice for most of applications.

From a comparison among stand-alone approaches in table 3, the following conclusions can be drawn:

- A very noticeable aspect is the great performance improving given by the use of deep learning: the three CNNs are the best stand-alone approaches in almost all the datasets and perform better than the fusion of all handcrafted methods (*Vote1*). For this reason they are weighed more than other approaches in other ensembles *Vote2*, *Vote3* and *Vote4* (*SegNet* has the weight of the sum of the handcrafted classifiers, *U-Net* has half the weight of *SegNet* and *DeepLab* has twice the weight of *SegNet*).
- In particular *DeepLab* is the best stand-alone method. The great advantage of *DeepLab* and *SegNet* with respect to *U-Net* is that the training of *SegNet* and *DeepLab* has been performed starting from a pre-trained model (VGG19 and ResNet50, respectively), while *U-Net* has been trained starting from random weights. This is a not negligible aspect considering that our training set contains only 2,000 images. The performance of the three networks can be certainly boost by a further training on a more specific dataset: i.e. classification in HGR may be improved by training using hand recognition images.
- The ensembles *Vote2* and *Vote3* have similar performance, even if *Vote3* outperforms *Vote2* (p-value 0.2305 using Wilcoxon signed rank test (McDonald, 2014)). Both the ensembles outperforms all the previous approaches (p-value di 0.05), excluding *DeepLab*. *Vote4*, which also includes *DeepLab* is the best method tested in our experiments.
- If we consider only the handcrafted methods, there is a noticeable difference among performance of different approaches in each dataset. The best stand-alone method is different for each dataset, anyway *SA2* and *SA3* work better than others do, this is the reason why we gave them the higher weights in our vote rule. In particular, *SA3* reaches the best performance (among handcrafted classifiers) in three datasets: *MCG*, *SFA*, *HGR*. In our opinion, it is a very valuable method in particular for problems where the background is quite simple (e.g. gesture recognition).
- The method *Bayes* performs very well (it is ranked third among handcrafted stand-alone methods) even though it is a very simple approach with low computational requirements.
- The statistical color model proposed by *Chen* is very efficient but at the expense of performance. Its low performance (last position in the rank) is certainly related to the absence of a tuning parameter.
- The method *SPL* performs very well in *VMD* but very poorly in *Compaq*. *SPL*, which is based on lookup tables, is a method suited for simple tasks, as gesture recognition, where the images are taken in fixed acquisition conditions and without complex background. This is the reason of its poor performance in *Compaq* which is a very difficult dataset.

For many approaches, the selection of an appropriate threshold is crucial for performance: the most appropriate choice depends on the specific application, since a strict value decreases the number of false positives despite of the true positive rate, vice versa a low threshold increases the true positive rate but also the number of false positives.

The method that we propose as the best solution is *DeepLab*; for this approach, besides $F_1$ measure, we also report *TPR* and *FPR* as performance indicators in the MGC and UChile datasets, for an easier comparison with several methods in the literature that use these two indicators in these datasets:

- *MGC* (*TPR*=0.832701, *FPR*=0.020698),
- *UChile* (*TPR*=0.876188, *FPR*=0.015276).

Visual comparisons are also presented in Fig. 1, 2 and 3, which show the results of the tested methods on three different samples taken in various illumination conditions and with different background. The three images are taken from the Schmugge dataset; from left-top to right-bottom input image, ground truth, and the probability masks of the following methods are shown: *GMM, Bayes, SPL, Cheddad, Chen, SA1, SA2, SA3, DYC, SegNet, U-Net, DeepLab*.

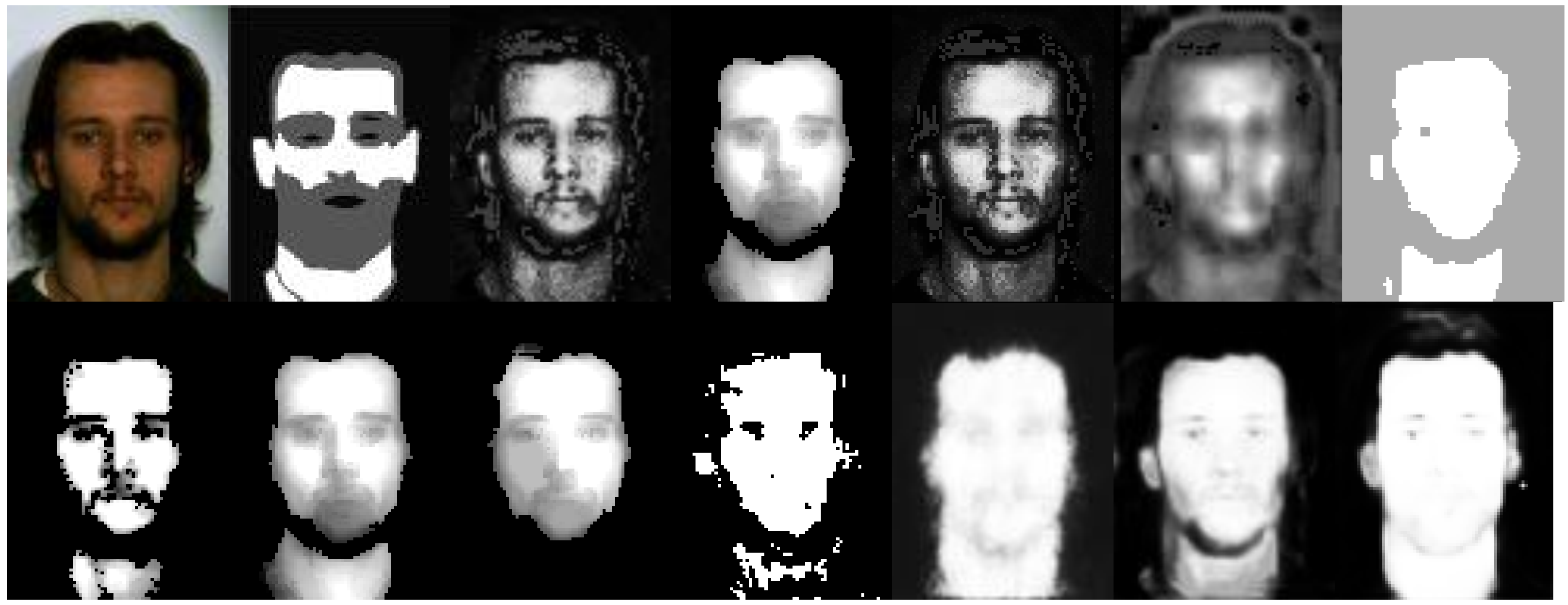

**Figure 1.** Schmugge005.jpg: (from left top to right bottom) input, ground truth, GMM, Bayes, SPL, Cheddad, Chen, SA1, SA2, SA3, DYC, SegNet, U-Net, DeepLab.

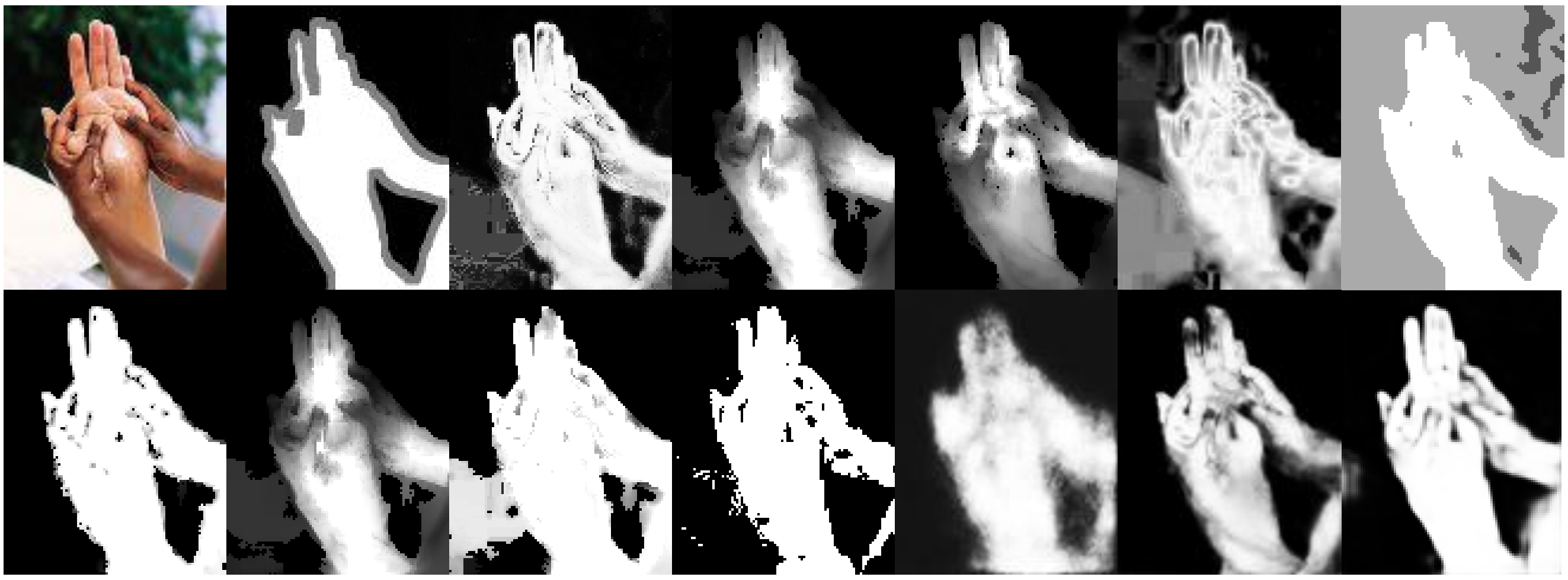

**Figure 2.** Schmugge367.jpg: (from left top to right bottom) input, ground truth, GMM, Bayes, SPL, Cheddad, Chen, SA1, SA2, SA3, DYC, SegNet, U-Net, DeepLab.

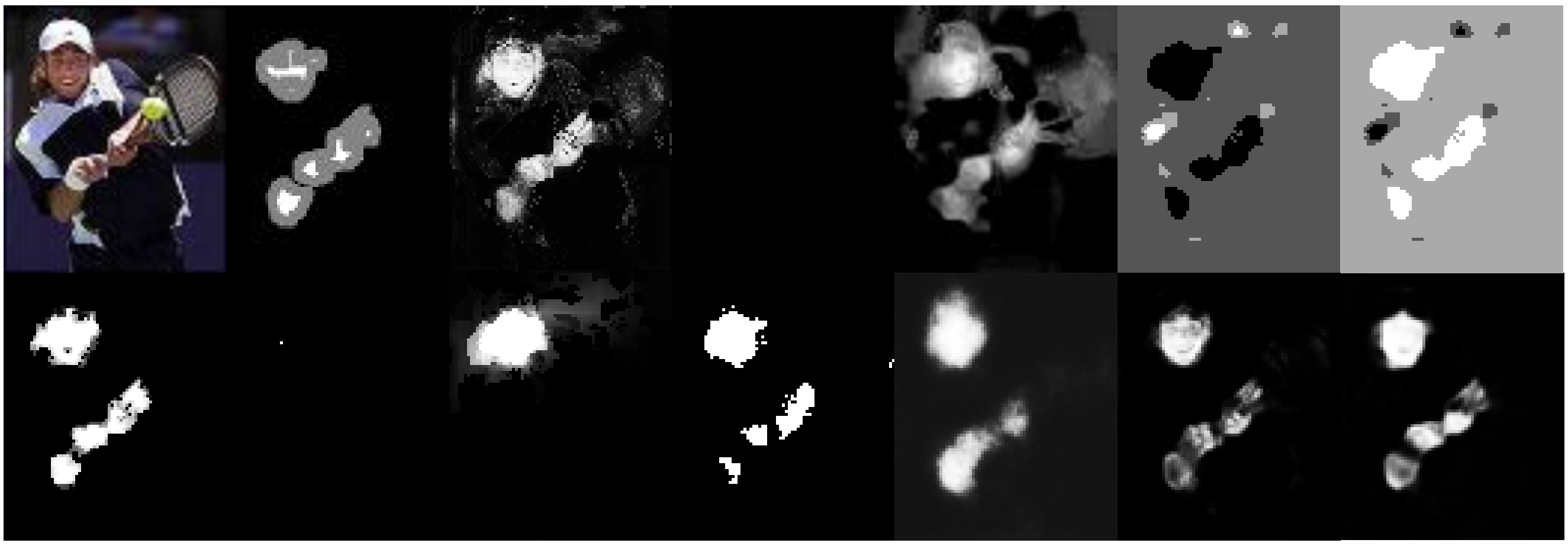

**Figure 3.** Schmugge845.jpg: (from left top to right bottom) input, ground truth, GMM, Bayes, SPL, Cheddad, Chen, SA1, SA2, SA3, DYC, SegNet, U-Net, DeepLab.

A deeper analysis of strengths and weaknesses of the solutions investigated can be carried out considering the characteristics of each approach summarized in Table 1, the performance results reported in Table 3 and the execution time reported in Table 4. The last column in Table 4 reports the execution time in seconds for mask calculation in a dataset of 100 images of size 640×480 (using non optimized MATLAB 2019 code on Intel Xeon CPU E5 1603 v4 @2.80GHz 64 GB RAM with NVIDIA TitanXp).

All the pixel-based approaches (i.e. *GMM*, *Bayes*, *SPL*, *Cheddad*, *Chen* and *DYC*) have the advantage of being fast, easy to implement and require low resources. Unfortunately, they are quite imprecise, mostly in detecting skin over complex background and when dealing with images very different from those used for training. These approaches can be used in application where skin detection is a preliminary filtering step and where there is a strong constraint on memory or CPU resources. The methods based on based on spatial analysis (i.e. *SA1*, *SA2*, *SA3*) are more complex than those of the previous class, require a higher computation (usually they start from the result of a pixel based method) but grant a better precision, mainly if trained on a similar dataset. The low performance in terms of execution time for this class of approaches is due to non-optimized and non-parallelized source code, therefore the comparison with CNN is not totally fair. Anyway, they were the best solution for application requiring high accuracy before the rise of Convolutional Neural Networks, but now they are outperformed by the deep learning approaches. The last group includes the deep learning approaches (i.e. *SegNet*, *U-Net*, *DeepLab*): these methods strongly outperform old approaches even if trained on a small dataset of only 2000 images and promise a further performance improvement if tuned on a specific problem. Ensembles deserve a separate discussion: they exploit the advantage of their components in terms of precision but at a cost of higher complexity (the execution time is not reported in Table 4 since it depends on the degree of parallelization of the code). The real advantage of a fusion approach is the possibility of combining methods trained on different datasets and get the best from each: this fact is paid in terms of increased complexity and computational requirements. As stated above, we think that a deep network is the best solution for all the application that requires high precision, otherwise we suggest a pixel-based method i.e. SPL which is fast and simple. A general evaluation of all methods compared in this work is summarized in Table 4 where pros and cons of each approach is given.

**Table 4. Final evaluation of the compared methods and execution time for 100 images.**

| Method | *Pros* | *Cons* | *Time (secs)* |
|---|---|---|---|
| *GMM* | Simple, fast: only for simple problems | Not precise for complex background | 4.11 |
| Bayes | Simple | Low accuracy | 14.56 |
| *SPL* | Fast: good for simple problems | Need parameter tuning | 2.86 |
| *Cheddad* | Fast | Need parameter tuning, imprecise | 3.30 |
| *Chen* | Very fast | Very imprecise | 2.64 |
| *SA1* | - | Need parameter tuning, imprecise | 24.68 |
| *SA2* | Quite precise | Need parameter tuning, very slow | 94.70 |
| *SA3* | Quite precise | Need parameter tuning, very slow | 142.97 |
| *DYC* | No parameter tuning | Quite imprecise | 23.45 |
| *SegNet* | Precise | High complexity (memory and CPU/GPU) | 9.36 |
| *U-Net* | Precise and fast | High complexity (memory and CPU/GPU) | 3.31 |
| *DeepLab* | Very precise | High complexity (memory and CPU/GPU) | 6.27 |

## 5. Conclusion and future research directions

In this work a new framework to evaluate and combine different skin detector approaches is presented and an extensive evaluation of several approaches is carried out on 10 different datasets including more than 10000 labelled images. A survey of most recent existing approaches is carried out, three well-known deep learning models for data segmentation are trained and tested to this classification problem and four new ensembles based on the combination of nine methods (including 3 CNNs) are proposed and evaluated. Experimental results confirm that CNNs work very well for skin segmentation and outperform all previous approaches based on handcrafted features; moreover, our proposed ensembles obtain a very good performance with respect to other stand-alone approaches. The performance obtained on 10 different datasets confirms that DeepLabv3+, which according to our experiments is the best performing stand-alone approach, is well-suited for detecting skin in a wide range of images, without requiring re-training and ad hoc parameter optimization and can be used in many skin detection applications such as body tracking, face detection, detection of objectionable content and hand gesture analysis. Anyway, considering the impact on the performance of the "deep learning revolution" we cannot avoid suggesting to start from a trained CNN and fine-tuning on the particular problem to have even more valuable performance.

In conclusion, we show that skin detection is a very difficult problem, which can be hardly solved by a stand-alone approach. The performance of many old methods for skin detection depended on many factors, such as the color space used, the parameters used, the nature of data, the image characteristics, the shape of the distribution, the size of samples for training, the presence of data noise, etc. New methods based on deep learning suffer less of these problems, mainly thanks to the possibility of training to very large samples (two of three CNNs tested in this work started from pretrained models). In any case a fine-tuning of the specific problem can help improving performance. Our experiments on ensembles show that fusing different approaches is a feasible solution for problems where training is impracticable since they are able to exploit the advantages of each composing method.

The recent deep learning revolution has carried out a rapid evolution in the field of image segmentation with several new model proposed in the last few years (A. Khan, Sohail, Zahoora, & Qureshi, 2020). The great advantage of the new deep learning approaches against traditional machine learning-based models is that the first only needs data and it does not need any traditional handcrafted feature engineering techniques. This make the same model suitable for many different applications (i.e. medical diagnosis, gesture recognition, autonomous car driving, object detection). Anyway, this is actually a big problem in the field of skin segmentation, because as shown in this survey, there is a lack of a "universal" dataset including different types/colors of skin that can be used to train a model. Therefore we suggest for future investigation, the collection and labelling of a large dataset including people from different regions of the world. Secondly further investigation should be addressed to develop lightweight architectures able to work on resource constraint hardware without compromising the performance.

As future work for designing an effective skin-color detector we consider to test a preprocessing module able to deal with the color constancy problem in order to improve the robustness to illumination changes and a postprocessing module based on morphological operators.

**Acknowledgments** We would like to acknowledge the support that NVIDIA provided us through the GPU Grant Program. We used a donated TitanX GPU to train CNNs used in this work.